\documentclass{bmcart}

\RequirePackage{natbib}
\usepackage[utf8]{inputenc} %
\usepackage{todonotes}
\usepackage{xargs}                      %
\reversemarginpar
\newcommandx{\todoluca}[2][1=]{\tikzexternaldisable\todo[linecolor=red,backgroundcolor=red!25,bordercolor=red,#1]{Luca: #2}\tikzexternalenable}
\newcommandx{\todopaolo}[2][1=]{\tikzexternaldisable\todo[linecolor=blue,backgroundcolor=blue!25,bordercolor=blue,inline,#1]{Paolo: #2}\tikzexternalenable}
\newcommandx{\todoelena}[2][1=]{\tikzexternaldisable\todo[linecolor=magenta,backgroundcolor=magenta!25,bordercolor=magenta,#1]{Elena: #2}\tikzexternalenable}

\usepackage{url}

\usepackage[linesnumbered,ruled,norelsize]{algorithm2e}

\usetikzlibrary{plotmarks}
\usepackage{pgfplots}
\usepgfplotslibrary{groupplots}

\usepackage{nameref}

\usepackage{textcomp}

\usepackage{subcaption}

\newcommand{\BACDef}{{\cal D}istributed {\cal A}ssociative {\cal C}lassifier} 
\newcommand{\BAC}{DAC} 

\def\Plus{\texttt{+}}
\def\Minus{\texttt{-}}
\usepackage{amssymb}

 \usepackage{multirow}
\usepackage{tikz-qtree}

\pgfplotsset{}

\startlocaldefs
\endlocaldefs

\begin{document}

\begin{frontmatter}

\begin{fmbox}
\dochead{Research}

\title{Scaling associative classification for very large datasets}

\author[
   addressref={aff1},                   %
   email={luca.venturini@polito.it}   %
]{\inits{L}\fnm{Luca} \snm{Venturini}}
\author[
   addressref={aff1},                   %
   email={elena.baralis@polito.it}   %
]{\inits{E}\fnm{Elena} \snm{Baralis}}
\author[
   addressref={aff1},                   %
   email={paolo.garza@polito.it}   %
]{\inits{P}\fnm{Paolo} \snm{Garza}}

\address[id=aff1]{%
  \orgname{Department of Control and Computer Engineering, Politecnico di Torino}, %
  \street{Corso Duca degli Abruzzi 24},                     %
  \city{Torino},                              %
  \cny{Italy}                                    %
}

\end{fmbox}%

\begin{abstractbox}

\begin{abstract} %
Supervised learning algorithms are nowadays successfully scaling up to datasets that are very large in volume, leveraging the potential of in-memory cluster-computing Big Data frameworks.
Still, massive datasets with a number of large-domain categorical features are a difficult challenge for any classifier.
Most off-the-shelf solutions cannot cope with this problem.

In this work we introduce \BAC , a \BACDef . \BAC\ exploits ensemble learning to distribute the training of an associative classifier among parallel workers and improve the final quality of the model.
Furthermore, it adopts several novel techniques to reach high scalability without sacrificing quality, among which a preventive pruning of classification rules in the extraction phase based on Gini impurity.
We ran experiments on Apache Spark, on a real large-scale dataset with more than 4 billion records and 800 million distinct categories. The results showed that \BAC\ improves on a state-of-the-art solution in both prediction quality and execution time.
Since the generated model is human-readable, it can not only classify new records, but also
allow understanding both the logic behind the prediction and the properties of the model,
becoming a useful aid for decision makers.
\end{abstract}


\end{abstractbox}

\end{frontmatter}

\section*{Introduction}

In the recent years, Big Data have received much attention by both the academic and the industrial world, with the aim of fully leveraging the power of the information they hide.
The dimensions on which very large datasets usually extend are mainly the size, i.e. the disk storage occupied, the volume, i.e. the number of records, the dimensionality, i.e. the number of features a record can have, and the domain, i.e. the number of distinct values a feature can take.
A special effort has been dedicated to Machine Learning algorithms, with a profusion of solutions to tackle the scalability problem, on some or all of the dimensions mentioned above.

Scalability on the domain dimension is a special concern for the datasets in which most of the features are categorical.
Categorical features have their values expressed in a discrete domain, and no concept of ordering or ranking can be assumed.
Discrete or discretized features are a special case of categorical features where an order among the values is defined.
The absence of a natural ordering increases the complexity of the treatment of categorical variables, as their values cannot be binned in groups or levels for example.

Associative classifiers are a special category of Machine Learning algorithms, where association rule mining is exploited for the purpose of classification.
In the past, they have proved to be able to produce classification models of high quality and outperform state-of-art algorithms like decision trees \cite{el2013comparison}.
Moreover, the model produced is readable, as it is made of association rules, can be debugged and even manually tuned if needed, by modifying or deleting specific rules.
In a world where the dimensions involved in a Machine Learning process go far beyond the human control, the ability to understand and tune the model created should not be underestimated.
The high readability can also foster a better understanding of the underlying processes and guide the decision-making operations towards effective actions.

Adapting an associative classifier to cope with very large data volumes has a number of obstacles.
Some of these are inherited from association rule and frequent itemsets mining, 
which usually extract a large set of association rules or frequent itemsets, sometimes larger than the dataset itself.
This behavior is clearly unsustainable with very large datasets, and state-of-art solutions strive to scale up to high-cardinality and high-dimensional data \cite{pulvirentisurvey}.
On the other hand, associative classifiers fit categorical domains particularly well \cite{THABTAH2007AMining},
and they have the potential to outperform state-of-the-art algorithms on this task. Works like \cite{bechini2016mapreduce:mrac,Venturini2016BAC:Frameworks} have already attempted to bring an Associative Classifier on a distributed computing framework, and proved the feasibility of such a system.

In this work, we propose a \BACDef, in short \BAC, which trains an ensemble model in a distributed computing framework.
Our reference architecture for the computing framework is an in-memory cluster computing framework like Apache Spark, on which we perform our experimental session.
To achieve high scalability without sacrificing quality, we adopt several novel solutions that effectively exploit the advantages of in-memory computing, like a greedy, preventive pruning of rules in the extraction phase based on Gini impurity, a model consolidation phase to produce a lightweight model and a majority voting scheme based on multiple rules.
We test our approach on a categorical dataset that is large in size (over 1TB), volume (more than 4 billion records) and domain (800 million distinct values among all the features).
We evaluate the quality and the performance reached against a state-of-the-art solution. %
The code of \BAC~is freely available as open source\footnote{ The code is publicly available at https://gitlab.com/dbdmg/dac}.

The article is organized as follows.
Section \nameref{sec:background} introduces the reader to important concepts behind associative classification.
Section \nameref{sec:approach} explains how \BAC~works, and Section \nameref{sec:experiments} describes the experimental evaluation of \BAC.
Section \nameref{sec:rw} provides an insight on the related previous literature.
Finally, Section \nameref{sec:conclusion} draws conclusions.

\section*{Background}
\label{sec:background}
In this section, we introduce the reader to a set of concepts specific of association rule mining and Associative Classifiers (ACs).
In association rule mining, a set of rules is automatically extracted from a dataset. The dataset is usually represented as a set of transactions, where each transaction is itself represented as a set of items, called itemset.
In associative classification, one of the items is the class item, or label.

Equivalently, in classification, the dataset is represented as a structured table of records and features.
Each feature is identified by a \emph{feature\_id}, that is set to some value $v$ for each record, or to a null value for not available information.
In this work, we will mainly refer to the first notation, but the mapping between the two is straightforward: a simple concatenation of \emph{feature\_id} and $v$ will serve as item for the transaction, or record.
Not available data are represented again with a null value, or not represented at all, as transactions do not have a fixed structure.
The common practice in classification is to define a training set, i.e. a part of the labeled dataset that is used to train the algorithm, and a test set, from which the labels are removed.
The two sets are used together with other techniques, like cross-validation, to simulate the behavior of the algorithm towards unlabeled, new data and validate its performance.

Association rules are made of an \emph{antecedent} itemset $A$, and a \emph{consequent} itemset $B$, and are read as $A$ yields $B$, or $A \Rightarrow B$.
When the consequent is made of a single item, and specifically an item belonging to the set of class labels, the association rule can be used to label the record. We inherit the naming in \cite{cba:ma1998integrating} and call these rules Class Association Rules, or CARs.

Both association rules and CARs share a number of metrics that measure their strength and statistical significance \cite{THABTAH2007AMining}.
We here list the ones mentioned in this paper.
The support count (\emph{supCount})
of an itemset is the number of transactions in the dataset $D$ that contain the whole itemset.
The support of a rule  $A \Rightarrow B$ is defined as $\emph{supCount}({A \cup B}) / |D|$, where $|D|$ is the cardinality of $D$.
The confidence of the rule is defined as  $\emph{supCount}({A \cup B}) / \emph{supCount}({A})$, and in CARs it measures how precise the rule is at labeling a record.
The $\chi^2$ of a CAR is the value of the $\chi^2$ statistics computed against the distribution of the classes in $D$, which states whether the assumption of correlation between the antecedent and the consequent is statistically significant.

Another measure that is widely used in classification algorithms is the Gini impurity \cite{breiman1996some}.
The Gini impurity measures how often a record would be wrongly labeled, if labeled randomly with the distribution of the classes in the dataset. 
It is used for example in decision trees, to evaluate the quality of the splits at each node.
Given N classes, the Gini impurity of a dataset, or portion of it, is computed as
$$ Gini = \sum_{i=1}^N{f_i(1-f_i)} $$
where $f_i$ is the frequency of class $i$ in the dataset, or portion of it, for which we are computing the impurity.
A portion of dataset is considered pure if its Gini is equal to 0, that happens when only a single label appears.
We will refer to the Gini Impurity of an itemset, as the impurity of the portion of the dataset that contains the itemset.

\section*{The proposed approach}
\label{sec:approach}

Traditionally, the training phase of an associative classifier is a memory-intensive process, often executed out-of-core.
The vast majority of the techniques has at least an instant of time where a very large set of itemsets or rules has been extracted and not yet pruned.
This model cannot leverage the advantages of our reference architecture, an in-memory cluster computing framework like Apache Spark.
In building a scalable associative classifier, we have been guided by the two following design principles:
i) anticipating pruning before the actual extraction of the rules, and ii) moving from a large model that predicts with only the first matching rule toward a lightweight model, that compensates the loss in size by applying all the rules that match.
These two principles aim at reducing the amount of rules contemporarily present in the main memory at any given instant of time, allowing for an effective exploitation of the in-memory computing platform.

The baseline framework on which we build for the training of our \BACDef, namely \BAC, is as follows.
\begin{enumerate}
\item The dataset is split into $N$ partitions, each one sampled from the original dataset with a ratio $r$;
\item Within each partition, a rule extraction phase occurs, that produces a model as a set of CARs.
The CARs found are filtered by minimum support, minimum confidence and minimum $\chi^2$ and optionally further pruned with a database coverage phase;
\item The generated $N$ models are collected in an ensemble.
\end{enumerate}

Following our first design principle, we aimed at devising an extraction phase that made the work of the posterior pruning extremely reduced or null, in the best case.
We have therefore adopted a greedy approach based on the Gini impurity of an item, keeping in mind the second design principle presented before, that we finally want a smaller model where several rules can collaborate for the prediction, instead of a single first-match.
This calls for shorter rules, that can more easily match new records and avoid over-fitting.
In order to follow such a route without sacrificing predictive quality we designed several solutions that will be presented in the next sections, namely: %
i) an FP-growth-like CAR extractor that produces only useful classification rules, in a greedy fashion, by exploiting the Gini impurity;
ii) an added model consolidation
phase for the generation of the ensemble that reduces further the size of the final model;
iii) new voting strategies for the ensemble that exploit the before-mentioned novelties.

\newcommand{\CARtree}{CAP-tree}
\newcommand{\CARtreeDef}{Class Association Patterns tree}
\newcommand{\CARgrowth}{CAP-growth}
\newcommand{\CARgrowthDef}{Class Association Patterns growth}
\subsection*{\CARgrowth}
The FP-tree is an effective solution for frequent itemsets extraction, and is often adapted to the extraction of CARs \cite{THABTAH2007AMining}.
Moreover, it adapts well to in-memory computing, as its construction needs only two scans of the dataset and, once built, the FP-tree stores in the main memory all the necessary information for frequent itemsets or CARs extraction.

However, there is a twofold motivation behind designing an alternative to the FP-tree, like \cite{li2001cmar,Baralis2008AClassification}, as method of storage for the patterns that will build the final CARs.
First, the FP-tree is designed to build all frequent patterns, that are a superset of what we look for when we build CARs.
Second, being frequent does not always coincide with being useful,
and using the standard FP-growth algorithm would lead the growth of an overwhelming number of rules that would impede the descent to lower supports, where more useful information may dwell.
Guided by these considerations, and keeping in mind the design principles outlined in the beginning of the section, we designed an FP-growth-like algorithm called \CARgrowth, for \CARgrowthDef.

\CARgrowth~stores the information that is useful for extracting CARs in a \CARtree.
Similarly to an FP-tree, this structure allows to compactly store all the information needed to extract association rules reading the dataset only twice.
Differently from the FP-tree, a \CARtree~stores in each node extra information useful to extract only CARs, as it is usually done in single-machine approaches\cite{li2001cmar,Baralis2008AClassification}.
Moreover, the first phase of the \CARtree's construction sorts the frequent items by their Gini impurity, which will help the extraction of more useful rules in the \CARgrowth~phase.

The algorithm that builds a \CARtree~is detailed in Algorithm \ref{alg:captree}.

  \begin{algorithm}[H]

	\label{alg:captree}
   	\caption{\CARtree~building}
	\SetKwInOut{Input}{Input}\SetKwInOut{Output}{Output}
    \SetKwProg{Fn}{Function}{}{}

	\Input{A transaction DB labeled with classes - $DB$}
	\Input{A minimum support threshold - $minsup$}
	\Output{A \CARtree}
	\BlankLine

	Scan the DB once.
    Collect $L$, the list of frequent items ($support >= minsup$).
    Sort $L$ by decreasing $IG$ and filter out items with $IG \leq 0$. \label{alg:captree:1pass}
    
    Create the root of a \CARtree~$T$ and label it as \emph{null}. \label{alg:captree:2pass}
    
    \For(){each labeled transaction $t$}
    {
    select only the items in $t$ that appear in $L$ and sort them according to the order in
    $L$, obtaining $t'$ \label{alg:captree:selectitems}
    
    call insert($t'$, $T$) \label{alg:captree:callinsert}
    } \label{alg:captree:2passend}
    
    \BlankLine
	\Fn{insert (transaction $t$, node $T$)}
    {
    $h$ = first item of $t$
    
    \eIf{$T$ has a child $T'$ s.t. $T'.id = h.id$ \label{alg:captree:updatestart}} 
    {
    $T'$.freqs[t.class]+=1
    }{
    create a new node $T'$
    
    init $T'$.id = h.id and $T'$.freqs to an array of zeros
    
    $T'$.freqs[$t$.class]+=1
    
    $T'$.parent = $T$
    
    update the header table \label{alg:captree:headertable}
   }\label{alg:captree:updateend}
    $t'$ = $t \backslash h$
    
    \If{$t'$ is not empty}{insert($t'$, $T'$)}
    
    }

  \end{algorithm}

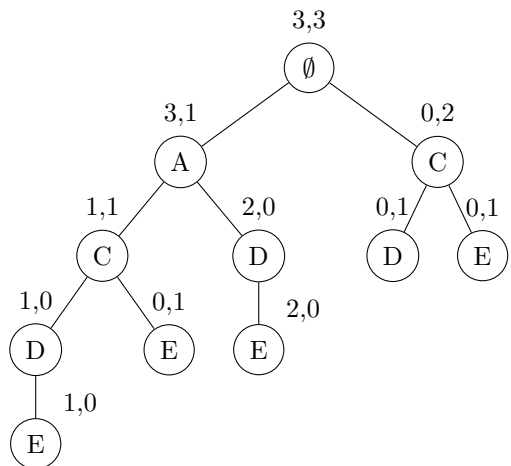
\begin{figure}
\caption{\csentence{A \CARtree~example}
      Built over the toy dataset with minimum support equal to 0.3}
\label{fig:toy_captree}

\begin{tikzpicture}[every tree node/.style={draw,circle},
   level distance=1.25cm,sibling distance=.5cm, 
   edge from parent path={(\tikzparentnode) -- (\tikzchildnode)}]
\Tree [.\node[label={3,3}] {$\emptyset$} ;
    [.\node[label={3,1}] {A} ;
      [.\node[label={1,1}] {C} ;
      	[.\node[label={1,0}] {D} ; 
        	[.\node[label=45:{1,0}] {E} ;]
        ]
        [.\node[label={0,1}] {E} ;]
      ] 
      [.\node[label={2,0}] {D} ;
      	[.\node[label=45:{2,0}] {E} ; ]
      ] 
    ]
    [.\node[label={0,2}] {C} ;
      [.\node[label={0,1}] {D} ;]
      [.\node[label={0,1}] {E} ;]
    ] ]
\end{tikzpicture}

\begin{tabular}{|l|l|l|}
\hline
\multicolumn{3}{|c|}{Header Table}                         \\ \hline
\multirow{2}{*}{Item}   & \multicolumn{2}{l|}{Freqs} \\ \cline{2-3} 
                        & +               & -              \\ \hline
A                       & 3               & 1              \\ \hline
C                       & 1               & 3              \\ \hline
D                       & 3               & 1              \\ \hline
E                       & 3               & 2              \\ \hline
\end{tabular}
\end{figure}

Given a minimum support threshold, which is used to recognize frequent itemsets, the algorithm scans the dataset twice.
In the first pass (line \ref{alg:captree:1pass}), it builds a list $L$ of frequent items, with decreasing and strictly positive Information Gain, which is computed as follows:
\begin{equation} \label{eq:ig}
 IG_i = Gini_D - [w_i Gini_i + (1-w_i) Gini_D]
\end{equation}

in which $Gini_D$ is the impurity of the global dataset, $Gini_i$ is the impurity of item $i$, and $w_i$ is the ratio of dataset containing the item.
Since we are considering the item alone, we assume that the $(1-w_i)$-th part of the dataset not covered by the item has a distribution of the labels identical to global distribution (same Gini).%

Equation \ref{eq:ig} simplifies as
\begin{equation} \label{eq:ig2}
IG_i = w_i (Gini_D - Gini_i)
\end{equation}

In this first passage, we can also obtain the frequency of the classes in the entire dataset, which is used in the \CARgrowth's extraction phase.
In the second pass (lines \ref{alg:captree:2pass}-\ref{alg:captree:2passend}), we insert each read transaction in the \CARtree~ (line \ref{alg:captree:callinsert}), maintaining a header table that keeps track of the pointers to the nodes in the tree that store the frequent items, like in the original FP-tree (line \ref{alg:captree:headertable}).
Before being inserted, the transaction is cleaned from the infrequent items and reordered according to the order of $L$ (decreasing IG) (line \ref{alg:captree:selectitems}).
The insertion updates the structure of the \CARtree~ to keep track of the label of the transaction in an array of frequencies (lines \ref{alg:captree:updatestart}-\ref{alg:captree:updateend}).
This allows the direct extraction of CARs and the computation of the IG and the confidence of the rules in the \CARgrowth.
Figure \ref{fig:toy_captree}   shows the \CARtree~built on the toy dataset in Table \ref{tab:ds_ex}, with the minimum support threshold set to 0.3, that is 2 records.
Each node of the tree is labeled with the array of the frequencies of the classes, positive and negative respectively.
In this tree we see how item $B$ has been pruned, since its IG is 0%
, and how the remaining items are sorted and inserted by their IG, with item $A$ being the first and the most useful for classification.

\CARgrowth~extracts a set of CARs from the \CARtree~descending the tree greedily.
This yields that, since the frequent items are sorted by decreasing IG, we evaluate the rules made of high-IG items first.
The rationale that guided the design of the algorithm is to avoid redundant rules, where possible, while keeping the length of the rules minimal.

The following example illustrates some ways in which redundancy affects CARs.
In other approaches, this redundancy is often reduced after the extraction of CARs, as shown in the \nameref{sec:rw}.
We provide this example so that the reader may later gain an intuition of where \CARgrowth~helps reducing redundancy before the extraction itself. 
In Figure \ref{fig:ex_cars} we see all the CARs in the set of association rules extracted with the standard FP-Growth, with minimum support set to 0.3 (2 rows or more) and minimum confidence 0.51 on the toy dataset in Table \ref{tab:ds_ex}.
18 CARs for a dataset of 6 records are clearly redundant.
A first, evident source of this redundancy is item $B$, which is present in all the records in the dataset.
This results in having, for any rule generated, an identical rule with $B$ appended, that does not contribute to the classification and lengthens the model.
A similar situation happens with item $E$.
Likewise, item $C$ appears in many rules, all of which agree in classifying a record as negative: item $C$ itself would be sufficient as antecedent of the rule.
The same holds for other rules as well.

\begin{table}[h]
\caption{An example transactional dataset, binary-labeled.}

\begin{center}
\begin{tabular}{ccc} 
\hline
\textbf{TID} & \textbf{Transaction} & \textbf{Class} \\ \hline
1	& $ \{A,B,D,E\} $	& + \\ \hline
2	& $ \{B,C,E\} $		& - \\ \hline
3	& $ \{A,B,D,E\} $	& + \\ \hline
4 	& $ \{A,B,C,E\} $	& - \\ \hline
5   & $ \{A,B,C,D,E\} $ 	& + \\ \hline
6 	& $ \{B,C,D\} $		& - \\ \hline
\end{tabular}
\end{center}
\label{tab:ds_ex}
\end{table}

\begin{figure}
\caption{An example model with CARs for the dataset in Table \ref{tab:ds_ex}}
\label{fig:ex_cars}

\begin{tabular}{cc}
$E D A \Rightarrow \Plus$ & $B E D A \Rightarrow \Plus$\\
$E D \Rightarrow \Plus $ & $B E D \Rightarrow \Plus $  \\
$E C \Rightarrow \Minus $  & $B E C \Rightarrow \Minus $    \\
$E A \Rightarrow \Plus $    &$B E A \Rightarrow \Plus $     \\
$D A \Rightarrow \Plus $    & $B D A \Rightarrow \Plus $     \\
$A \Rightarrow \Plus $    &  $B A \Rightarrow \Plus $ \\ 
$E \Rightarrow \Plus $ & $B E \Rightarrow \Plus $\\
$C \Rightarrow \Minus $ & $B C \Rightarrow \Minus $ \\
$D \Rightarrow \Plus$ & $B D \Rightarrow \Plus$

\end{tabular}

\end{figure}

As previously stated, \CARgrowth~aims to avoid the redundancy of the example above.
Algorithm \ref{alg:capgrowth} shows the pseudocode for \CARgrowth.
Similarly to Equation \ref{eq:ig2}, we define the Information Gain for a node as
\begin{equation} \label{eq:ig3}
IG_T = w_T (Gini_{T.parent} - Gini_T)
\end{equation}
where $w_T$ is the ratio of transactions represented in node $T$ with regards to its parent node, and Gini is computed on the frequencies of the labels stored in the node.

  \begin{algorithm}[h]
   	\caption{\CARgrowth}
    \label{alg:capgrowth}
	\SetKwInOut{Input}{Input}\SetKwInOut{Output}{Output}
    \SetKwProg{Fn}{Function}{}{}

	\Input{a \CARtree~}
    \Input{A minimum support threshold - $\textrm{minsup}$}
	\Input{A minimum confidence threshold - $\textrm{minconf}$}
    \Input{A minimum chi2 threshold - $\textrm{minchi2}$}
	\Output{A list of CARs}
	\BlankLine
    rules = $\emptyset$

    \For{each child $T$ of \CARtree.root}{

    rules += extract($T$)
    }
    
    return rules
    \BlankLine
	\Fn{extract(node T) \label{alg:capgrowth:extract}}
    {
    rules = $\emptyset$
    
    \If(//negative Information Gain: do not generate any rule){$IG(T) <= 0$}{
    
    return $\emptyset$ \label{alg:capgrowth:ig0}}
    \If(//pure node: try to generate a rule){$Gini(T) == 0$}{return generateRule($T$) \label{alg:capgrowth:gini0}}
    \For{each child $T'$ of $T$ \label{alg:capgrowth:childstart}}
    {
    rules += extract($T'$)
    }
    \If(//none of the children has produced a rule: try to generate a rule){rules is $\emptyset$}
    {
    return generateRule($T$)
    }\label{alg:capgrowth:childend}
    return rules
    }
    \BlankLine
    \Fn{generateRule(node T) \label{alg:capgrowth:generate}}
    {
    consequent = class with highest value in $T$.freqs[]
    
    antecedent = set of items in the path from $T$ to \CARtree.root

	tree = \CARtree~conditioned by the items in antecedent \label{alg:capgrowth:captree}

    freqs = tree.root.freqs \label{alg:capgrowth:freqs}
    
    sup = freqs[consequent] / totCount \label{alg:capgrowth:sup}
    
    supAntecedent = freqs.sum / totCount
    
    from sup, supAntecedent and the global frequencies of the classes computed in the first pass of Algorithm \ref{alg:captree} compute support, confidence and $\chi^2$ for the generated rule: antecedent $\Rightarrow$ consequent \label{alg:capgrowth:computerule}
    
    \If{$\textrm{sup} < \textrm{minsup}$ or $\textrm{conf} < \textrm{minconf}$ or $\chi^2 < \textrm{minchi2}$ \label{alg:capgrowth:thresholds}}{return $\emptyset$}
    return rule \label{alg:capgrowth:end}
    }
  \end{algorithm}
  
  \begin{figure*}
      \caption{\csentence{\CARgrowth~}
      Example visit of the \CARtree~in Figure \ref{fig:toy_captree}}
        \label{fig:capgrowth_running}
\begin{tabular}{cc}
\begin{subfigure}{\linewidth/2}

  \begin{tikzpicture}[every tree node/.style={draw,circle},
   level distance=1.25cm,sibling distance=.5cm, 
   edge from parent path={(\tikzparentnode) -- (\tikzchildnode)}]
\Tree [.\node[label={3,3 0.5}] {$\emptyset$} ;
\edge node[right, auto=left] {IG: 8.3\%};
    [.\node[label=-90:{3,1 0.375}] {A}  ;
     ]]
\end{tikzpicture}
\caption{}
\label{fig:capgrowth1}
\end{subfigure}

  &
\begin{subfigure}{\linewidth/2}
  \begin{tikzpicture}[every tree node/.style={draw,circle},
   level distance=1.25cm,sibling distance=.5cm, 
   edge from parent path={(\tikzparentnode) -- (\tikzchildnode)}]
\Tree [.\node[label={3,3 0.5}] {$\emptyset$} ;
    [.\node[label=30:{3,1 0.375}] {A} ;
      \edge node[right, auto=left] {IG: -6.25\%};
      [.\node[label=-90:{1,1 0.5}] {C} ;
      	]]]
\end{tikzpicture}
\caption{}
\label{fig:capgrowth2}
\end{subfigure}
  \\
\begin{subfigure}{\linewidth/2}
  \begin{tikzpicture}[every tree node/.style={draw,circle},
   level distance=1.25cm,sibling distance=.5cm, 
   edge from parent path={(\tikzparentnode) -- (\tikzchildnode)}]
\Tree [.\node[label={3,3 0.5}] {$\emptyset$} ;
    [.\node[label=30:{3,1 0.375}] {A} ;
      [.\node[label=-90:{1,1 0.5}] {C} ;
      ] 
      \edge node[auto=left] {IG: 18.75\%};
      [.\node[label=-90:{2,0 0.0}] {D} ;
      ]]]
      	
\end{tikzpicture}
\caption{}
     \label{fig:capgrowth3}
\end{subfigure}

  &
\begin{subfigure}{\linewidth/2}

\begin{tikzpicture}[every tree node/.style={draw,circle},
   level distance=1.25cm,sibling distance=.5cm, 
   edge from parent path={(\tikzparentnode) -- (\tikzchildnode)}]
\Tree [.\node[label={3,3 0.5}] {$\emptyset$} ;
    [.\node[label=100:{3,1 0.375}] {A} ;
      [.\node[label=-90:{1,1 0.5}] {C} ;
      ] 
      [.\node[label=-90:{2,0 0.0}] {D} ;
      ] 
    ]
    \edge node[auto=left] {IG: 16.6\%};
    [.\node[label=-90:{0,2 0.0}] {C} ;
    ] ]
\end{tikzpicture}
\caption{}
    \label{fig:capgrowth4}
\end{subfigure}
\end{tabular}
  \end{figure*}

The algorithm is a recursive call to the function extract (line \ref{alg:capgrowth:extract}), which visits in a depth-first fashion the \CARtree.
The stopping criteria of this visit are:
\begin{enumerate}
\item a negative Information Gain for the current node.
In this case, we do not generate any rule (line \ref{alg:capgrowth:ig0}).
\item a Gini impurity for the current node equal to 0. Being the Gini impurity always strictly decreasing, this makes the current node the first pure node in the path from the root to this node, i.e. we see only one label for it.
We try to generate a rule(line \ref{alg:capgrowth:gini0}).
\end{enumerate}
Whenever none of the children of a node does generate a rule, the node itself tries to generate a new rule (lines \ref{alg:capgrowth:childstart}-\ref{alg:capgrowth:childend}).
This can occur when the children nodes do not see enough samples to satisfy the minimum support threshold, for example, or if the current node is a leaf.

The function that generates a new rule (lines \ref{alg:capgrowth:generate}-\ref{alg:capgrowth:end}) needs first to recollect the frequencies of the labels from all the nodes where the current pattern appears.
Like in the original FP-growth, this is done by projecting the \CARtree~recursively on all the items of the pattern, that is all the nodes in the path to the root (line \ref{alg:capgrowth:captree}).
At the end of the projection, the root node contains the array of classes' frequencies for the pattern (line \ref{alg:capgrowth:freqs}).
With it, we can compute the support, the confidence and the $\chi^2$ of the rule we are trying to generate (lines \ref{alg:capgrowth:sup}-\ref{alg:capgrowth:computerule}).
If any of the measures does not satisfy the minimum constraints, the rule is not generated (line \ref{alg:capgrowth:thresholds}).

In Figure \ref{fig:capgrowth_running}
we see an example of the \CARgrowth~algorithm, run on the \CARtree~of Figure \ref{fig:toy_captree}, with minimum support, confidence and $\chi^2$ set respectively to 0.3, 0.51 and 0.
In the figure, each node is labeled by the array of frequencies of the classes and the resulting Gini impurity.
The root of the tree has a Gini impurity of 0.5.
Its first child to be explored stores item $A$ with a Gini of 0.375 (Figure \ref{fig:capgrowth1}).
Having a positive IG and a non-null Gini, we continue the descent to its children.
The first to be explored describes the pattern ${A,C}$ (Figure \ref{fig:capgrowth2}).
This node has a Gini index of 0.5, thus a negative IG.
This means that the addition of item $C$ to the pattern only worsens the ability of $A$ in predicting a label.
We therefore do not explore anymore this pattern and its offsprings.
The other sibling (Figure \ref{fig:capgrowth3}), storing item $D$, is pure for the positive class: continuing the descent further would only lengthen the rule without any improvement.
We reconstruct the real frequencies of itemset $\{A, D\}$ to see if the rule $A, D \Rightarrow \Plus$ is really worth to generate and compute its support, confidence, and $\chi^2$.
First, we need to project the \CARtree~for item $D$.
The header table stores the pointers to the three nodes that store this item.
Only the parts of the tree that end to these three nodes are kept, and all the surviving nodes and the header table are updated in their frequency arrays to reflect this change.
Now we have a \CARtree~storing only the transactions that contain item $D$.
We project again for item $A$.
The header table points to a single node that stores this item, and its frequency array, updated in the step before, is [3,0].
By projecting, the \CARtree~reduces to the root node alone, whose frequencies also are updated to [3,0].
This is the frequency array for itemset $\{A, D\}$.
Thus, the rule  $A, D \Rightarrow \Plus$ has confidence 1 and support 0.5, and satisfies the minimum thresholds\footnote{The minimum $\chi^2$ is set to 0 in this example.}.
Rule $A \Rightarrow \Plus$ is not generated, as one of the subpatterns of $A$ has already produced one rule.
Finally, we move to the second child of the root, storing item $C$ (Figure \ref{fig:capgrowth4}).
This is again a pure node.
We recollect the frequencies of item $C$ by projection as seen before and get the array [1,3], which produces the rule $C \Rightarrow \Minus$ with support 0.5 and confidence 0.75.
The final model is made of only two rules.

It is worth paralleling the strategy in \CARgrowth~with the one in the database coverage pruning \cite{cba:ma1998integrating}.
The database coverage scans the rules extracted and already sorted by prediction quality, and keeps on adding them to the model if they predict correctly at least a transaction not yet covered, and until all the transactions have been covered at least once.
Similarly, \CARgrowth~keeps on adding rules that cover transactions not yet covered, since they are extracted in different branches of the \CARtree, and does so without extracting the entire set of CARs that satisfy the minimum thresholds.
The main difference between the two strategies is in the moment when the pruning is performed: the database coverage acts at the end of the extraction, when all the rules have been already extracted, whereas \CARgrowth~anticipates the pruning in the extraction phase.
The aim of both the strategies is the same, that is generating the least, shortest rules, avoiding redundancy in the model.

\subsection*{Model consolidation}
 \CARgrowth~generates a single model, in each partition of the dataset, that is at the same time compact and useful.
Still, with massively large datasets, it may happen that the number of partitions to have a sufficient division of the workload is in the order of thousands, or more.
Consequently, the number of single models in the ensemble explodes.
This results in a larger model to store, more complex to be read and examined by a human, and with longer execution times when applied to predict new records.

To cope with these issues, we shrink the ensemble of the models to a unique model.
This is done by merging the models, combining rules with identical antecedent and consequent into a single, new rule.
The new rule will need to have an approximation for its support, confidence and $\chi^2$, as it is too expensive to reconstruct the exact ones in this phase.
In other words, we anticipate part of the voting that eventually classifies new records to this phase: establishing how two identical rules collapse to a single one is establishing how they would eventually vote in the classification, a priori.
Algorithm \ref{alg:compression} shows how the consolidation is done.

  \begin{algorithm}[H]

	\label{alg:compression}
   	\caption{Model consolidation}
	\SetKwInOut{Input}{Input}\SetKwInOut{Output}{Output}
    \SetKwProg{Fn}{Function}{}{}

	\Input{A list of models - $models$}
	\Output{A single model, as a list of CARs}
	\BlankLine

model = $\emptyset$

\For{each m in models}{
model = merge(model, m) \label{alg:line:reduce}
}

return model

\BlankLine

\Fn{merge(model m1, model m2)}{
m = new model

rules = m1.rules $\cup$ m2.rules \label{alg:line:union}

gr = group rules by same antecedent and consequent

\For{each group of rules $i$ in gr}{
m = m $\cup$ aggregate($i$) \label{alg:line:aggr}
}

return m
}

\Fn{aggregate(rules)}{
rule = new rule

r = rules.first

(rule.antecedent, rule.consequent) = (r.antecedent, r.consequent)

supports = $\bigcup_{r \in \emph{rules}}$r.support

confs = $\bigcup_{r \in \emph{rules}}$r.confidence

chis = $\bigcup_{r \in \emph{rules}}$r.chi2

(rule.support, rule.confidence, rule.chi2) = \emph{g}(supports, confs, chis) \label{alg:line:somefun}

return rule
}

\Fn{\emph{g}(supports, confs, chis)}{
return (max(supports), max(confs), max(chis)) \label{alg:compression:g}

}

  \end{algorithm}

We recall that \BAC~'s training has split the dataset in $N$ partitions and runs a \CARgrowth~over each partition, thus generating an ensemble of $N$ models.
These models are the input for the model consolidation algorithm (Algorithm \ref{alg:compression}).
We reduce the models by applying, recursively two by two, a function \emph{merge} (line \ref{alg:line:reduce}).
This function simply makes the union of the rules in the two models (line \ref{alg:line:union}) and, for each set of identical (in the antecedent and consequent) rules found, applies function \emph{aggregate} (line \ref{alg:line:aggr}). 

Function \emph{aggregate} returns a new rule by choosing the new support, confidence and $\chi^2$ with \emph{g()} 
 (line \ref{alg:line:somefun}), which actually sets the strategy for the consolidation.
The default behavior of $g()$ is returning the maximum of the supports, confidences and $\chi^2$ in input, as an upper bound estimation (line \ref{alg:compression:g}).
We have also experimented with other possibilities, namely functions that keep the property of associativity and commutativity, i.e. the minimum and the product.
Associativity and commutativity in function $g()$ make the consolidation runnable in parallel.
In Section \nameref{sec:dacparams} we give details on these experiments.
\subsection*{Voting}
\label{sec:voting}
In associative classifiers, the models usually label a record by applying the first matching rule based on a quality ranking.
Differently from other families of classifiers, associative classifiers usually do not have a score 
or a vector of probabilities for the prediction, but only the predicted class.
Introducing a score for the prediction of the associative classifier, the predictions can express their strength in a continuous domain and we can use measures different from the accuracy to compare the model with others, like the Area Under Curve. 
Moreover, we have a way to weight the votes in the ensemble, whereas in its simplest implementation every model would have voted with an equal weight, independently of the confidence or support of the rules of each model.
This last point is indeed partially covered by the consolidation, but we can still hold in the consolidated model rules, with different antecedents or consequents, that come from different models and contemporarily match a record.
Defining a score would mean defining how these many rules contribute to strengthen our belief in predicting a class, when they all agree, or to mitigate our certainty, when they partially disagree.

Given an unlabeled record, for each label $i$, we define a score $p_i$ as a function of some measure for all rules matching the record, i.e.
$$p_i = f(m(\vec{r_i})),\qquad \forall{i : \vec{r_i} \not= \emptyset} $$
where $\vec{r_i}$ is the array of matching rules for label $i$, $m$ is a measure, e.g. the support or the confidence, and $f$ a function 
with domain in $[0,1]$.
If there are no matching rules for a label and the record, $p_i$ is defined as
$$p_i = p_X/|X|,\qquad \forall{i \in X} $$
where $X$ is set of labels for which we do not have a matching rule, and $p_X$ is defined under a naive assumption of independence as
$$p_X = \prod_{j: \vec{r_j} \not= \emptyset}{(1-p_j)} ,\qquad  X = \{i : \vec{r_i} = \emptyset\}$$

If there are no matching rules at all, $p_i$ is default to the probability of each label $i$ in the original dataset.
The score vector $\vec{p}$, containing the scores $p_i$ as above defined, is finally normalized to sum to one.

The default setting for $m$ is the confidence, that is a common choice in associative classifiers for the rules' ranking.
In preliminary experiments, we tried several alternative choices for $m$, i.e. the support, its complement ($1-support$) and the $\chi^2$.
We performed further experiments on the two most promising of these, that is the confidence and $1-support$, which we report in Section \nameref{sec:dacparams}.

The default setting for $f()$ is the $max()$ function, which is an upper bound estimation of the quality of the rule, based on the measures from the models where it was found.
Alternatives to this choice are, for example, the minimum or the mean, which are always valid scores whenever $m(\vec{r_i})$ is defined between 0 and 1.
We test and discuss these alternatives in Section \nameref{sec:dacparams}.

\section*{Experimental evaluation}
\label{sec:experiments}
In our experimental evaluation, we want to compare \BAC~with state-of-art approaches in a realistic, large-scale scenario.
Among publicly available datasets, we found only one dataset to be very large (i.e. over the Terabyte) and with the characteristics of our problem (i.e. many categorical features), and is described below.
As competitors to \BAC, we choose the algorithms implemented in the Apache Spark Mllib library \cite{mllib2016meng}, as it is a well-proven framework for machine learning on distributed computing \cite{landset2015survey,singh2015survey}.

The experiments were performed on a cluster with 30 worker nodes 
running Cloudera Distribution of Apache Hadoop (CDH5.8.2), 
which comes with Spark 1.6.0.
The cluster has 2TB of RAM, 324 cores, and 773TB of secondary memory. 
Unless differently specified, all the single experiments are run on 100 executors and a master node with one virtual core and 7GB of RAM each.
We used version 1.6.0 of Apache Spark Mllib\footnote{https://spark.apache.org/docs/1.6.0/mllib-guide.html} and version 2.1 of \BAC, which is released as open source\footnote{https://gitlab.com/dbdmg/dac/tags/v2.1}.

The dataset used in the experiments is the Criteo dataset \cite{Chapelle2014SimpleAdvertising}, which has already been used as a benchmark in classification tasks, although only on its continuous features, in \cite{chen2016xgboost}.
The dataset counts more than 4 billion records, describing the behavior of users in 24 consecutive days towards web ads.
The positive class is a click on the showed ad and the negative is a non-click.
The records are described by 13 continuous features and 26 categorical features, whose semantics is not disclosed.
For the experiments, we selected the categorical features only, as \BAC~does not handle continuous features without a discretization phase, which is outside the scope of this evaluation.
The resulting dataset contains more than 800 million unique items, each appearing once or more, and is larger than 1.2 TB. 
The negative class appears 97\% of the times.

The dataset is characterized by the presence of categorical features and the extreme imbalance of the classes.
In the next paragraphs, we will describe our approach toward the two issues.

\textbf{Managing categorical features.} To deal with categorical features, we need either an algorithm that supports them natively or a proper encoding of the features into integer or binary values.
A common solution, which would enable the exploitation of many widely-used classification algorithms, like SVMs or artificial neural networks, is to use the so-called ``one-hot'' encoding.
With it, all the distinct values appearing in the dataset are transformed to a binary feature, which represents the presence or absence of the value in the record.
With all the categorical features mapped to binary ones, we would be able to try many solutions for classification.

We tried one-hot encoding as implemented in Mllib.
Unfortunately, with so many unique values (more than 800 million) the preprocessing quickly grows in memory and fails.
A possible reason is the fact that the records are stored in a dense vector.
Since with this encoding only a few features would be non-zero, we tried to implement the encoding with a sparse matrix, but the dimensions involved (billions of records by billions of features) showed to be too large also for this kind of representation, and our attempts exhausted the memory available to our testbed.

A different approach is selecting an algorithm that supports natively categorical features without a special encoding, like decision trees or random forests.
Again, the number of distinct values in each feature is an issue, due to the metadata that these algorithms need to collect and store to decide the binnings and the splits at each iteration.
Not surprisingly, all preliminary experiments again failed for out-of-memory errors.
We decided therefore to exploit a technique known as ``hashing trick'' \cite{WeinbergerFeatureLearning}.
With this method, all values are hashed to reduce dimensionality, with inevitable collisions.
We therefore progressively reduced the domain of each feature down to 100000 categories%
, value that allowed the execution of the random forest algorithm without memory issues.
After the reduction of dimensionality, the application of one-hot encoding was still impossible.
This therefore excludes from our analysis the Mllib implementations of linear SVM, logistic regression and multilayer perceptron.

\textbf{Dealing with class imbalance.} Preliminary experiments showed that neither Random Forests nor \BAC~were able to handle the highly unbalanced distribution of classes in this dataset.
Indeed, the resulting models were respectively trees with all the leaf nodes predicting the majority class and sets of CARs where the minority class was highly underrepresented, when not absent.
To cope with this issue, we investigated several techniques, among which instance-based weighting, oversampling, and subsampling.
Instance-based weighting assigns a given weight $w$ to each sample, that while building the model is thus counted as if present $w$ times.
In the decision tree and the random forest, this weight affects the sample counts of each node and the split decisions.
When the weight $w$ is equal to the inverse of the frequencies of the sample's class, this technique can balance the dataset without a physical replication of the records.
Although implemented in several popular random forest implementations \cite{scikit-learn,r-randomforest}, instance-based weighting is not implemented in Mllib.
Oversampling replicates some of the records belonging to the minority class or classes, so that the dataset gets balanced \cite{witten2016data}.
In our scenario, the application of this technique to the training set did not converge successfully due to memory constraints.
Conversely, subsampling extracts a fraction of the majority class or classes, to reduce their volume to a size comparable to the minority class \cite{witten2016data}.
We applied this technique to the negative class to have a cardinality roughly equal to the positive one, in the training set.
The test set was not subsampled.

In these preliminary evaluations we have also tried several settings of the general architecture of \BAC.
In this phase, we found that sampling with replacement yields a better load balancing, as this operation triggers the shuffling of part of the training dataset and leads to equally-sized partitions, whereas the default partition can see blocks of very different sizes.
We have set to $1/N$ the sampling size for each one of the $N$ models, to have a final training dataset sized as the original one.
We have also tried several $N$, finding in 100 for each partition a value that allowed the \CARtree~of each model to be stored in memory.

Summarizing, our competitor to \BAC~will be a Random Forest with categorical features, with the values of the categories hashed down to 100000 different values at most\footnote{Hashing to larger values or not using hashing was not a viable option for the memory issues explained before.}.
\BAC~will be instead evaluated without hashing trick, as it is not necessary.
For both, we will subsample the majority class in the training set.

\subsection*{Experimental comparison of \BAC~and Random Forest}

In this section, we evaluate the quality and the performance of \BAC~and a Random Forest.
Our objective is to show how our proposed technique can manage a dataset characterized by a very large volume and domain, and compare the quality of the resulting model with the state of the art.
We evaluate our results in the binary tasks with the AUROC, i.e. the area under the ROC curve \cite{bradley1997auroc}.
\BAC~will be evaluated with its default settings, i.e. $f() = max$, $m =$ confidence and $g() = max$. 
The experiments are run with a 5-fold cross-validation on the whole dataset, with the K-fold function implemented in the MLUtils of Mllib.
Each one had a variable duration on our testbed between 2 and 30 hours.
All the confidence intervals shown in the plots were computed using a t-student distribution at 95\% confidence.

\begin{figure}[h!]

\caption{\csentence{\BAC~vs Random Forest.}
      AUROC cross-validated with 5-fold on the entire dataset}
 \label{fig:auroc}     
\begin{tikzpicture}

\begin{axis}[
ylabel={AUROC},
xmin=0, xmax=10,
ymin=0.61, ymax=0.66,
xtick={1,2,3,4,5,6,7,8,9},
xticklabel style={rotate=90},
xticklabels={Decision
Tree,Random Forest
5-tree,Random Forest
10-tree,Random Forest
100-tree,\BAC~0.01,\BAC~0.005,\BAC~0.002,\BAC~0.001,\BAC~0.0002},
tick align=outside,
xmajorgrids,
x grid style={lightgray!92.026143790849673!black},
ymajorgrids,
y grid style={lightgray!92.026143790849673!black}
]
\path [draw=blue, semithick] (axis cs:1,0.615497649528459)
--(axis cs:1,0.61637188419107);

\path [draw=blue, semithick] (axis cs:2,0.633516262853185)
--(axis cs:2,0.63531139184836);

\path [draw=blue, semithick] (axis cs:3,0.637652851496438)
--(axis cs:3,0.638214831818147);

\path [draw=blue, semithick] (axis cs:4,0.643078698841512)
--(axis cs:4,0.64570005524175);

\path [draw=red, semithick] (axis cs:5,0.628827448881859)
--(axis cs:5,0.639426247169403);

\path [draw=red, semithick] (axis cs:6,0.638795299789913)
--(axis cs:6,0.640710414993112);

\path [draw=red, semithick] (axis cs:7,0.643351003653274)
--(axis cs:7,0.644548476855172);

\path [draw=red, semithick] (axis cs:8,0.646350573496082)
--(axis cs:8,0.648048028011691);

\path [draw=red, semithick] (axis cs:9,0.655011191405508)
--(axis cs:9,0.656841213747655);

\addplot [semithick, blue, mark=*, mark size=3, mark options={solid,fill=white}]
table {%
1 0.615934766859765
};
\addplot [semithick, blue, mark=pentagon*, mark size=3, mark options={solid,fill=white}]
table {%
2 0.634413827350773
};
\addplot [semithick, blue, mark=triangle*, mark size=3, mark options={solid,fill=white}]
table {%
3 0.637933841657292
};
\addplot [semithick, blue, mark=triangle*, mark size=3, mark options={fill=white,solid,rotate=180}]
table {%
4 0.644389377041631
};
\addplot [semithick, red, mark=*, mark size=3, mark options={solid}]
table {%
5 0.634126848025631
};
\addplot [semithick, red, mark=triangle*, mark size=3, mark options={solid}]
table {%
6 0.639752857391513
};
\addplot [semithick, red, mark=triangle*, mark size=3, mark options={solid,rotate=90}]
table {%
7 0.643949740254223
};
\addplot [semithick, red, mark=square*, mark size=3, mark options={solid}]
table {%
8 0.647199300753887
};
\addplot [semithick, red, mark=diamond*, mark size=3, mark options={solid}]
table {%
9 0.655926202576582
};
\end{axis}

\end{tikzpicture}      
\end{figure}
     
        \begin{figure}[h!]
  \caption{\csentence{\BAC~vs Random Forest}
      AUROC vs training and testing times, cross-validated with 5-fold on the entire dataset}
      
\label{fig:auroc_time}
      
\begin{tabular}{cc}
   
\begin{subfigure}{\linewidth}
\centering
\begin{tikzpicture}[font=\footnotesize]

\begin{axis}[
xlabel={training time (s)},
ylabel={AUROC},
xmin=0, xmax=18900,
ymin=0.61, ymax=0.66,
tick align=outside,
xmajorgrids,
x grid style={lightgray!92.026143790849673!black},
ymajorgrids,
y grid style={lightgray!92.026143790849673!black},
legend style={at={(0.97,0.03)}, anchor=south east, draw=white!80.0!black},
legend cell align={left},
legend entries={{Decision Tree},{Random Forest 5-tree},{Random Forest 10-tree},{Random Forest 100-tree},{\BAC~0.01},{\BAC~0.005},{\BAC~0.002},{\BAC~0.001},{\BAC~0.0002}}
]
\path [draw=blue, semithick] (axis cs:3846.61154137222,0.615934766859765)
--(axis cs:4089.32400558058,0.615934766859765);

\path [draw=blue, semithick] (axis cs:3967.9677734764,0.615497649528459)
--(axis cs:3967.9677734764,0.61637188419107);

\path [draw=blue, semithick] (axis cs:2435.89076934057,0.634413827350773)
--(axis cs:2588.02618499543,0.634413827350773);

\path [draw=blue, semithick] (axis cs:2511.958477168,0.633516262853185)
--(axis cs:2511.958477168,0.63531139184836);

\path [draw=blue, semithick] (axis cs:3080.07082290903,0.637933841657292)
--(axis cs:3234.09449229257,0.637933841657292);

\path [draw=blue, semithick] (axis cs:3157.0826576008,0.637652851496438)
--(axis cs:3157.0826576008,0.638214831818147);

\path [draw=blue, semithick] (axis cs:15383.66756152,0.644389377041631)
--(axis cs:18578.5369099232,0.644389377041631);

\path [draw=blue, semithick] (axis cs:16981.1022357216,0.643078698841512)
--(axis cs:16981.1022357216,0.64570005524175);

\path [draw=red, semithick] (axis cs:416.307080791489,0.634126848025631)
--(axis cs:460.050111488511,0.634126848025631);

\path [draw=red, semithick] (axis cs:438.17859614,0.628827448881859)
--(axis cs:438.17859614,0.639426247169403);

\path [draw=red, semithick] (axis cs:414.791173405449,0.639752857391513)
--(axis cs:460.724957377351,0.639752857391513);

\path [draw=red, semithick] (axis cs:437.7580653914,0.638795299789913)
--(axis cs:437.7580653914,0.640710414993112);

\path [draw=red, semithick] (axis cs:419.88177226605,0.643949740254223)
--(axis cs:467.90451035755,0.643949740254223);

\path [draw=red, semithick] (axis cs:443.8931413118,0.643351003653274)
--(axis cs:443.8931413118,0.644548476855172);

\path [draw=red, semithick] (axis cs:423.484319342598,0.647199300753887)
--(axis cs:465.570963389002,0.647199300753887);

\path [draw=red, semithick] (axis cs:444.5276413658,0.646350573496082)
--(axis cs:444.5276413658,0.648048028011691);

\path [draw=red, semithick] (axis cs:475.682192063964,0.655926202576582)
--(axis cs:526.810233439636,0.655926202576582);

\path [draw=red, semithick] (axis cs:501.2462127518,0.655011191405508)
--(axis cs:501.2462127518,0.656841213747655);

\addplot [semithick, blue, mark=*, mark size=3, mark options={solid,fill=white}]
table {%
3967.9677734764 0.615934766859765
};
\addplot [semithick, blue, mark=pentagon*, mark size=3, mark options={solid,fill=white}]
table {%
2511.958477168 0.634413827350773
};
\addplot [semithick, blue, mark=triangle*, mark size=3, mark options={solid,fill=white}]
table {%
3157.0826576008 0.637933841657292
};
\addplot [semithick, blue, mark=triangle*, mark size=3, mark options={fill=white,solid,rotate=180}]
table {%
16981.1022357216 0.644389377041631
};
\addplot [semithick, red, mark=*, mark size=3, mark options={solid}]
table {%
438.17859614 0.634126848025631
};
\addplot [semithick, red, mark=triangle*, mark size=3, mark options={solid}]
table {%
437.7580653914 0.639752857391513
};
\addplot [semithick, red, mark=triangle*, mark size=3, mark options={solid,rotate=90}]
table {%
443.8931413118 0.643949740254223
};
\addplot [semithick, red, mark=square*, mark size=3, mark options={solid}]
table {%
444.5276413658 0.647199300753887
};
\addplot [semithick, red, mark=diamond*, mark size=3, mark options={solid}]
table {%
501.2462127518 0.655926202576582
};
\end{axis}

\end{tikzpicture}
\caption{}
\label{fig:auroc_time:train}

\end{subfigure}

\\

\begin{subfigure}{\linewidth}
\centering
\begin{tikzpicture}[font=\footnotesize]

\begin{axis}[
xlabel={testing time/record (${\mu}s$)},
ylabel={AUROC},
xmin=0, xmax=18,
ymin=0.61, ymax=0.66,
tick align=outside,
xmajorgrids,
x grid style={lightgray!92.026143790849673!black},
ymajorgrids,
y grid style={lightgray!92.026143790849673!black},
legend style={at={(0.97,0.03)}, anchor=south east, draw=white!80.0!black},
legend cell align={left},
legend entries={{Decision Tree},{Random Forest 5-tree},{Random Forest 10-tree},{Random Forest 100-tree},{\BAC~0.01},{\BAC~0.005},{\BAC~0.002},{\BAC~0.001},{\BAC~0.0002}}
]
\path [draw=blue, semithick] (axis cs:0.820834341324104,0.615934766859765)
--(axis cs:0.834927078096498,0.615934766859765);

\path [draw=blue, semithick] (axis cs:0.827880709710301,0.615497649528459)
--(axis cs:0.827880709710301,0.61637188419107);

\path [draw=blue, semithick] (axis cs:0.931134926117939,0.634413827350773)
--(axis cs:0.978086151109425,0.634413827350773);

\path [draw=blue, semithick] (axis cs:0.954610538613682,0.633516262853185)
--(axis cs:0.954610538613682,0.63531139184836);

\path [draw=blue, semithick] (axis cs:1.23021616258981,0.637933841657292)
--(axis cs:1.26103773776512,0.637933841657292);

\path [draw=blue, semithick] (axis cs:1.24562695017747,0.637652851496438)
--(axis cs:1.24562695017747,0.638214831818147);

\path [draw=blue, semithick] (axis cs:3.03773793021952,0.644389377041631)
--(axis cs:3.19254836524026,0.644389377041631);

\path [draw=blue, semithick] (axis cs:3.11514314772989,0.643078698841512)
--(axis cs:3.11514314772989,0.64570005524175);

\path [draw=red, semithick] (axis cs:0.571095315304945,0.634126848025631)
--(axis cs:0.718573263273121,0.634126848025631);

\path [draw=red, semithick] (axis cs:0.644834289289033,0.628827448881859)
--(axis cs:0.644834289289033,0.639426247169403);

\path [draw=red, semithick] (axis cs:0.969457863532382,0.639752857391513)
--(axis cs:1.23456073377102,0.639752857391513);

\path [draw=red, semithick] (axis cs:1.1020092986517,0.638795299789913)
--(axis cs:1.1020092986517,0.640710414993112);

\path [draw=red, semithick] (axis cs:1.82248543230106,0.643949740254223)
--(axis cs:2.74714838876484,0.643949740254223);

\path [draw=red, semithick] (axis cs:2.28481691053295,0.643351003653274)
--(axis cs:2.28481691053295,0.644548476855172);

\path [draw=red, semithick] (axis cs:3.24218515046506,0.647199300753887)
--(axis cs:4.8228921190146,0.647199300753887);

\path [draw=red, semithick] (axis cs:4.03253863473983,0.646350573496082)
--(axis cs:4.03253863473983,0.648048028011691);

\path [draw=red, semithick] (axis cs:16.0715469413896,0.655926202576582)
--(axis cs:17.5505856014229,0.655926202576582);

\path [draw=red, semithick] (axis cs:16.8110662714062,0.655011191405508)
--(axis cs:16.8110662714062,0.656841213747655);

\addplot [semithick, blue, mark=*, mark size=3, mark options={solid, fill=white}]
table {%
0.827880709710301 0.615934766859765
};
\addplot [semithick, blue, mark=pentagon*, mark size=3, mark options={solid,fill=white}]
table {%
0.954610538613682 0.634413827350773
};
\addplot [semithick, blue, mark=triangle*, mark size=3, mark options={solid, fill=white}]
table {%
1.24562695017747 0.637933841657292
};
\addplot [semithick, blue, mark=triangle*, mark size=3, mark options={fill=white,solid,rotate=180}]
table {%
3.11514314772989 0.644389377041631
};
\addplot [semithick, red, mark=*, mark size=3, mark options={}]
table {%
0.644834289289033 0.634126848025631
};
\addplot [semithick, red, mark=triangle*, mark size=3, mark options={solid}]
table {%
1.1020092986517 0.639752857391513
};
\addplot [semithick, red, mark=triangle*, mark size=3, mark options={solid,rotate=90}]
table {%
2.28481691053295 0.643949740254223
};
\addplot [semithick, red, mark=square*, mark size=3, mark options={solid}]
table {%
4.03253863473983 0.647199300753887
};
\addplot [semithick, red, mark=diamond*, mark size=3, mark options={solid}]
table {%
16.8110662714062 0.655926202576582
};
\end{axis}

\end{tikzpicture}
\caption{}
\label{fig:auroc_time:test}
\end{subfigure}

\end{tabular}   
      
      \end{figure}

Figure \ref{fig:auroc} shows the resulting AUROC for the candidate set of models, consisting of
i) \BAC~with $f()=max$, $g()=max$ and $m=$ confidence, varying minimum support thresholds from 0.01 to 0.0002;
ii) Random Forests with a depth of 4, varying the number of trees from 5 to 100;
iii) a single Decision Tree, with depth 4.
The baseline for the results is set by the Decision Tree, which is almost two points below the Random Forest and \BAC.
The quite large confidence interval for \BAC~with 0.01 as minimum support makes uncertain the comparison with the two smallest forests, with 5 and 10 models respectively.
These last three models are all clearly below the results of the 100-tree forest and \BAC~with minimum support 0.002, that have a comparable AUROC of 0.644.
Significantly better are the results of \BAC~with minimum supports of 0.001 and 0.0002, this last one scoring the highest AUROC of 0.655, a good point above the 100-tree forest.

Notably, the experiments for the 100-tree forest lasted 30 hours in our testbed, against the 20 hours of the \BAC~with minimum support of 0.0002.
These high computation costs would certainly be a heavy factor in the choice of a model, as the model with the highest score is not always a viable path.
We therefore plot the same scores against the training and testing time of their models, in Figure \ref{fig:auroc_time}.
In Figure \ref{fig:auroc_time:train} we see how the training times of the Random Forest grow with the number of models.
The Decision Tree shows times higher than both the 5-tree and 10-tree forests, as it does not perform any feature selection, whereas the forests randomly choose $\sqrt{n}$ features for each tree, where $n$ is number of columns, i.e. 26 in our scenario.
The half-point advantage in the AUROC of the 100-tree forest on the 10-tree one comes with a cost five times higher in terms of training times.
This large gap could make the difference in a scenario where the model needs to be frequently updated, e.g. an application with nightly updates to the training dataset, and could lead to the choice of the shallower model.
\BAC~here demonstrates a highly desirable behavior, as the best model trains in only 500 seconds, a time respectively 5 and 25 times smaller than the 5-tree and the 100-tree forests, which also have a worse AUROC.
Moreover, the gap between the training times of the least and most accurate models for \BAC~is under the 15\%, so the latter one is clearly preferable.

The testing times of \BAC~ and the Random Forest have similar trends, depicted in Figure \ref{fig:auroc_time:test}. 
Both appear to grow exponentially with the AUROC reached, symptom of models that are more and more complex.
For the Random Forest, this complexity is proportional to the total number of splits, or equivalently to the number of trees, since we have a fixed depth.
This justifies the alignment of the Decision Tree to the trend of the Random Forests, as in this phase it is practically identical to a Random Forest with one tree.
For \BAC, the complexity depends on the number of rules extracted and, in this strategy that applies $max$ for both $f()$ and $g()$, on the position of the first matching rules in the model for each class, for we need only these for the score.
This explains the slightly larger confidence interval on the time axes, whereas the forests all have negligible intervals, due to the constant number of splits traversed by each record for a prediction.
Despite this, we can still safely affirm that \BAC~reaches the same quality of a Random Forest within smaller testing times, as it happens with \BAC~with minimum support 0.002 and the 100-tree forest.
At the same time, we can say that, given a comparable testing time, \BAC~performs better, as in the case with minimum support 0.005 and the 10-tree forest.

\subsection*{Evaluation of \BAC~parameters}
\label{sec:dacparams}
We tested the effect of the choice of the algorithms' parameters on the quality of the model, to eventually select one or several candidates for more thorough tests.

For \BAC, we evaluated different choices for:
\begin{itemize}
\item the use of the database coverage technique,\hfill [yes/no]
\item the function used in the voting, $f()$,\hfill [max/min/mean]
\item the measure used in $f()$, $m$,\hfill [confidence/1-support]
\item the function used in the model consolidation phase, $g()$,\hfill [max/min/product]
\item the minimum support threshold,\hfill [9 values from 5\% to 0.01\%]

\end{itemize}

for a total of 324 runs, considering all the combinations of values.
$f()$ was chosen among $max$, $min$ and $mean$. 
We tested two values for measure $m$, that is the confidence of the matching rules, which is a common choice in associative classifiers, and $1 - support$, following the intuition that a rule (a set of words) is the better in labeling the more is rare~\cite{Sebastiani:2002:MLA:505282.505283}.
$g()$ was chosen among $min$, $max$ and $product$, three functions that have the properties of associativity and commutativity, which are important for the distribution of the workload.
We tried nine different values for the minimum support threshold, from 5\% to 0.01\%.
The database coverage was either used or not.
The minimum confidence has been set to 50\%, for the rationale that any rule better than random guessing should positively contribute to the quality of the labeling.
The minimum $\chi^2$ was set to 3.841, corresponding to a p-value of 0.05 for the statistics.

We ran this session of experiments on the day 0 of the dataset, which is a 24th of the whole dataset, and without cross-validation, keeping 30\% of the dataset out for testing.
This reduced the execution time by more than two orders of magnitude, allowing us to test a larger selection of parameter values within some days of execution.

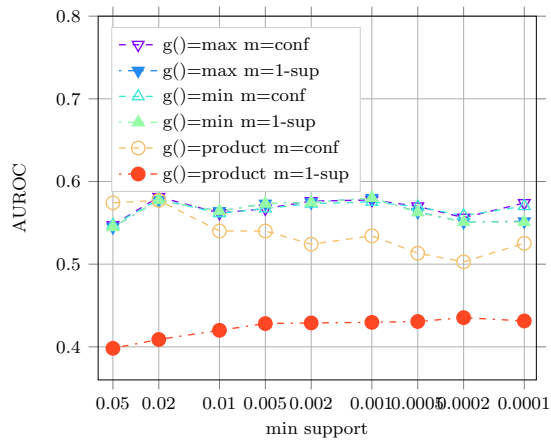
\begin{figure}[!ht]
\caption{\csentence{\BAC~tuning} Comparison of different choices for $f$, $g$, $m$ and \emph{minsup}}
\label{fig:bac_tuning}

\begin{tabular}{cc}

\begin{subfigure}{\linewidth}
\centering
\begin{tikzpicture}[scale=0.85, font=\footnotesize]

\definecolor{color1}{rgb}{0.13921568627451,0.53686659764418,0.960121645374628}
\definecolor{color0}{rgb}{0.5,0,1}
\definecolor{color3}{rgb}{0.590196078431372,0.989980213280707,0.655283850013454}
\definecolor{color2}{rgb}{0.229411764705882,0.911022649246088,0.840344071637893}
\definecolor{color5}{rgb}{1,0.279582592596744,0.141206151823091}
\definecolor{color4}{rgb}{0.958823529411765,0.751331889556873,0.412356317473904}

\begin{axis}[
xlabel={min support},
ylabel={AUROC},
xmin=8e-05, xmax=0.06,
ymin=0.36, ymax=0.8,
xmode=log,
xtick={0.05,0.02,0.01,0.005,0.002,0.001,0.0005,0.0002,0.0001},
xticklabels={0.0001,0.0002,0.0005,0.001,0.002,0.005,0.01,0.02,0.05},
tick align=outside,
xmajorgrids,
x grid style={lightgray!92.026143790849673!black},
ymajorgrids,
y grid style={lightgray!92.026143790849673!black},
legend entries={{g()=max
m=conf},{g()=max
m=1-sup},{g()=min
m=conf},{g()=min
m=1-sup},{g()=product
m=conf},{g()=product
m=1-sup}},
legend cell align={left},
legend style={at={(0.03,0.97)}, anchor=north west, draw=white!80.0!black}
]
\addplot [semithick, color0, dashed, mark=triangle, mark size=3, mark options={solid,rotate=180}]
table {%
0.0001 0.547162644108295
0.0002 0.581350754497667
0.0005 0.561785781534198
0.001 0.567201640751825
0.002 0.576226310988329
0.005 0.57769180405215
0.01 0.569921524462976
0.02 0.556291977943305
0.05 0.57392746272778
};
\addplot [semithick, color1, dash pattern=on 1pt off 3pt on 3pt off 3pt, mark=triangle*, mark size=3, mark options={solid,rotate=180}]
table {%
0.0001 0.544734629811027
0.0002 0.577200494944895
0.0005 0.564117126211464
0.001 0.573178505747828
0.002 0.574787468418062
0.005 0.580057358456742
0.01 0.563117856053315
0.02 0.550966381465807
0.05 0.551756771299847
};
\addplot [semithick, color2, dashed, mark=triangle, mark size=3, mark options={solid}]
table {%
0.0001 0.547162644108295
0.0002 0.578404362899988
0.0005 0.561600779351171
0.001 0.567321410368493
0.002 0.573131074121459
0.005 0.575431766384169
0.01 0.569407211128961
0.02 0.558694275376015
0.05 0.570213156393753
};
\addplot [semithick, color3, dash pattern=on 1pt off 3pt on 3pt off 3pt, mark=triangle*, mark size=3, mark options={solid}]
table {%
0.0001 0.544734629811027
0.0002 0.577200494944895
0.0005 0.56412649276763
0.001 0.573183804792638
0.002 0.574775017769691
0.005 0.580017930326918
0.01 0.562746812345492
0.02 0.551023329626778
0.05 0.550887678861008
};
\addplot [semithick, color4, dashed, mark=o, mark size=3, mark options={solid}]
table {%
0.0001 0.574219518322167
0.0002 0.577174121894512
0.0005 0.54011742724221
0.001 0.540095715673893
0.002 0.524140746522658
0.005 0.534228976544601
0.01 0.513235557897407
0.02 0.502838600256741
0.05 0.525253622563491
};
\addplot [semithick, color5, dash pattern=on 1pt off 3pt on 3pt off 3pt, mark=*, mark size=3, mark options={solid}]
table {%
0.0001 0.398238666310242
0.0002 0.408911674820009
0.0005 0.41999396565292
0.001 0.428109306152198
0.002 0.42890739277336
0.005 0.429627711615054
0.01 0.43064160447453
0.02 0.435290772936152
0.05 0.431309536578819
};
\end{axis}

\end{tikzpicture}
\caption{f() = min}
\label{fig:bac_tuning:min}
\end{subfigure}
\\
\begin{subfigure}{\linewidth}
\centering
\begin{tikzpicture}[scale=0.85, font=\footnotesize]

\definecolor{color1}{rgb}{0.13921568627451,0.53686659764418,0.960121645374628}
\definecolor{color0}{rgb}{0.5,0,1}
\definecolor{color3}{rgb}{0.590196078431372,0.989980213280707,0.655283850013454}
\definecolor{color2}{rgb}{0.229411764705882,0.911022649246088,0.840344071637893}
\definecolor{color5}{rgb}{1,0.279582592596744,0.141206151823091}
\definecolor{color4}{rgb}{0.958823529411765,0.751331889556873,0.412356317473904}

\begin{axis}[
xlabel={min support},
ylabel={AUROC},
xmin=8e-05, xmax=0.06,
ymin=0.36, ymax=0.66,
xmode=log,
xtick={0.05,0.02,0.01,0.005,0.002,0.001,0.0005,0.0002,0.0001},
xticklabels={0.0001,0.0002,0.0005,0.001,0.002,0.005,0.01,0.02,0.05},
tick align=outside,
xmajorgrids,
x grid style={lightgray!92.026143790849673!black},
ymajorgrids,
y grid style={lightgray!92.026143790849673!black},
legend style={at={(0.97,0.03)}, anchor=south east, draw=white!80.0!black},
legend entries={{g()=max
m=conf},{g()=max
m=1-sup},{g()=min
m=conf},{g()=min
m=1-sup},{g()=product
m=conf},{g()=product
m=1-sup}},
legend cell align={left}
]
\addplot [semithick, color0, dashed, mark=triangle, mark size=3, mark options={solid,rotate=180}]
table {%
0.0001 0.550071611717466
0.0002 0.604852694311312
0.0005 0.604335841018423
0.001 0.621811163807813
0.002 0.632469053410701
0.005 0.644975082166342
0.01 0.642223189278508
0.02 0.646857060933556
0.05 0.651571766612515
};
\addplot [semithick, color1, dash pattern=on 1pt off 3pt on 3pt off 3pt, mark=triangle*, mark size=3, mark options={rotate=180,solid}]
table {%
0.0001 0.547715032207451
0.0002 0.601435827986266
0.0005 0.599462468808145
0.001 0.612090460390452
0.002 0.619406906538084
0.005 0.62707228477099
0.01 0.624548808817636
0.02 0.623706646450276
0.05 0.628266721974667
};
\addplot [semithick, color2, dashed, mark=triangle, mark size=3, mark options={solid}]
table {%
0.0001 0.550071611717466
0.0002 0.604963476724233
0.0005 0.603587268843558
0.001 0.621510386736953
0.002 0.631708296670785
0.005 0.644252099414082
0.01 0.641423640952432
0.02 0.645606621691742
0.05 0.650783731980465
};
\addplot [semithick, color3, dash pattern=on 1pt off 3pt on 3pt off 3pt, mark=triangle*, mark size=3, mark options={solid}]
table {%
0.0001 0.547715032207451
0.0002 0.601433675638251
0.0005 0.599532073484231
0.001 0.61230938629032
0.002 0.619243468308968
0.005 0.62741961346252
0.01 0.62464405535457
0.02 0.623698848536273
0.05 0.62837654825779
};
\addplot [semithick, color4, dashed, mark=o, mark size=3, mark options={solid}]
table {%
0.0001 0.600317261383927
0.0002 0.600504303172485
0.0005 0.582055460690339
0.001 0.587256510876766
0.002 0.59518749516712
0.005 0.601746540524612
0.01 0.598948107582313
0.02 0.599664494203141
0.05 0.605322413366555
};
\addplot [semithick, color5, dash pattern=on 1pt off 3pt on 3pt off 3pt, mark=*, mark size=3, mark options={solid}]
table {%
0.0001 0.402729783201441
0.0002 0.534136768225155
0.0005 0.553907134095903
0.001 0.577675072165005
0.002 0.606932651124748
0.005 0.615112699743391
0.01 0.620164521098139
0.02 0.618837250626941
0.05 0.624965278311417
};
\end{axis}

\end{tikzpicture}
\caption{f() = max}
\label{fig:bac_tuning:max}

\end{subfigure}

\\

\begin{subfigure}{\linewidth}
\centering
\begin{tikzpicture}[scale=0.85, font=\footnotesize]

\definecolor{color1}{rgb}{0.13921568627451,0.53686659764418,0.960121645374628}
\definecolor{color0}{rgb}{0.5,0,1}
\definecolor{color3}{rgb}{0.590196078431372,0.989980213280707,0.655283850013454}
\definecolor{color2}{rgb}{0.229411764705882,0.911022649246088,0.840344071637893}
\definecolor{color5}{rgb}{1,0.279582592596744,0.141206151823091}
\definecolor{color4}{rgb}{0.958823529411765,0.751331889556873,0.412356317473904}

\begin{axis}[
xlabel={min support},
ylabel={AUROC},
xmin=8e-05, xmax=0.06,
ymin=0.36, ymax=0.66,
xmode=log,
xtick={0.05,0.02,0.01,0.005,0.002,0.001,0.0005,0.0002,0.0001},
xticklabels={0.0001,0.0002,0.0005,0.001,0.002,0.005,0.01,0.02,0.05},
tick align=outside,
xmajorgrids,
x grid style={lightgray!92.026143790849673!black},
ymajorgrids,
y grid style={lightgray!92.026143790849673!black},
legend style={at={(0.97,0.03)}, anchor=south east, draw=white!80.0!black},
legend entries={{g()=max
m=conf},{g()=max
m=1-sup},{g()=min
m=conf},{g()=min
m=1-sup},{g()=product
m=conf},{g()=product
m=1-sup}},
legend cell align={left}
]
\addplot [semithick, color0, dashed, mark=triangle, mark size=3, mark options={solid,rotate=180}]
table {%
0.0001 0.549403481802073
0.0002 0.600173829904405
0.0005 0.597645473702892
0.001 0.614181991205001
0.002 0.628233604610958
0.005 0.640121519607123
0.01 0.639196258160847
0.02 0.642188876828345
0.05 0.647908792617605
};
\addplot [semithick, color1, dash pattern=on 1pt off 3pt on 3pt off 3pt, mark=triangle*, mark size=3, mark options={solid,rotate=180}]
table {%
0.0001 0.546888771579272
0.0002 0.595493089677926
0.0005 0.589119420522399
0.001 0.604449634305801
0.002 0.609397400790683
0.005 0.615852392958573
0.01 0.608737647068609
0.02 0.615107566848944
0.05 0.61560207404524
};
\addplot [semithick, color2, dashed, mark=triangle, mark size=3, mark options={solid}]
table {%
0.0001 0.549502399754793
0.0002 0.600897350275687
0.0005 0.596887562621141
0.001 0.614042397143513
0.002 0.627661920525996
0.005 0.639419075072582
0.01 0.637877410235981
0.02 0.640502999437012
0.05 0.646807718020129
};
\addplot [semithick, color3, dash pattern=on 1pt off 3pt on 3pt off 3pt, mark=triangle*, mark size=3, mark options={solid}]
table {%
0.0001 0.546888771579272
0.0002 0.59551923260588
0.0005 0.589317529757649
0.001 0.604611685013246
0.002 0.596643174766321
0.005 0.616136950066755
0.01 0.608920432749681
0.02 0.615113205894324
0.05 0.615776780992559
};
\addplot [semithick, color4, dashed, mark=o, mark size=3, mark options={solid}]
table {%
0.0001 0.57492616252285
0.0002 0.590978653328013
0.0005 0.5640094122903
0.001 0.570465488390672
0.002 0.586313147426716
0.005 0.579549468091697
0.01 0.570936780181439
0.02 0.586815790837934
0.05 0.590953776268978
};
\addplot [semithick, color5, dash pattern=on 1pt off 3pt on 3pt off 3pt, mark=*, mark size=3, mark options={solid}]
table {%
0.0001 0.40192232058667
0.0002 0.448275172706943
0.0005 0.473788754162743
0.001 0.500850609825105
0.002 0.550272436844354
0.005 0.562619770192208
0.01 0.587439942027496
0.02 0.60299509791982
0.05 0.599178082084229
};
\end{axis}

\end{tikzpicture}
\caption{f() = mean}
\label{fig:bac_tuning:mean}

\end{subfigure}
\end{tabular}
\end{figure}

\textbf{Database coverage.} The first, immediate finding was on the use of the database coverage, which did not show effects on the quality of the model trained.
The amount of rules pruned by this technique has been constantly below 5\%. Thus,  \CARgrowth~is effective in selecting useful rules with limited overlapping.
For example, with a minimum support of 0.1\%, the number of rules of the model produced by \BAC~was 339, reduced to 328 with the database coverage.
The training time with this technique grew instead with the number of rules, motivating us not to use it in the following experiments.

Figure \ref{fig:bac_tuning}  shows the results of the runs without the database coverage.

\textbf{Function \boldmath$f()$.} Choosing $min$ as $f()$ (Figure \ref{fig:bac_tuning:min}) is comparable with other options only with shallow models (min sup 5\%).
With this support, the number of rules extracted (6) is so little that $f()$ rarely affects the voting.
Decreasing the support, $min$ does not show improvements, as only more confident and rare rules are being added to the models. Thus, the minimum $m$ does not change.
Both $f()=mean$ (Figure \ref{fig:bac_tuning:mean}) and $f()=max$ (Figure \ref{fig:bac_tuning:max}), instead, improve their performance with a similar rate, with the top performers almost overlapping, and the top AUROC for $max$ standing $0.4\%$ above the one for $mean$.%

\textbf{Measure \boldmath$m()$.} Preliminary experiments  already led us not to choose the support itself and the $\chi^2$ for $m()$.
Confidence proves to be the best choice.
Against the trend is the case where $g()$, in the consolidation function, is set to be the product of the measures.
In this case $1-support$ is the better choice for $m$, reaching an AUROC of 0.625 with $max$ as $f()$, ranking third among all experiments but still two points below the best scenario.

\textbf{Function \boldmath$g()$.} As for what concerns $g()$, the function applied to two identical rules in the consolidation phase to choose the new confidence, support and $\chi^2$, choosing either $min$ or $max$ is identical in this set of experiments.
Choosing $product$, instead, shows contrasting outcomes. %
Together with the confidence as $m$, it never shows improvements with lower supports, reaching at most an AUROC of 60\%.
With $f()$ set to $min$ (Figure \ref{fig:bac_tuning:min}), it has the worst quality among all the combinations, often below the AUROC of a random choice (50\%).
With $f()$ set to $max$ (Figure \ref{fig:bac_tuning:max}) and $1-sup$ as $m$, instead, as said above, it reaches the first quartiles of the results and is able to equal the AUROC of the alternatives at the lowest support.

\textbf{Minimum support.} With varying minimum support thresholds, from 0.02 to 0.0001, the best performing solution is stably with $f()=max$ (Figure \ref{fig:bac_tuning:max}) and $m=$ confidence, and indifferently $max$ or $min$ as $g()$.
This, with an arbitrary choice of $g=max()$, is the solution we tested on the whole dataset and compared with the state of the art.

\subsection*{Model selection for Random Forest}
With a Random Forest, the parameters that would affect the quality and the performance of the resulting model are mainly two, the number of trees and their depth.
Similarly to the previous section, we run some preliminary tests to evaluate different choices for these parameters, on the same portion of the dataset, again without cross-validation.

We evaluated the AUROC for depths of 4, 8 and 16, starting with 10 trees and increasing their number until possible.

\begin{figure}[h!]
  \caption{\csentence{Random Forest tuning.}
      Performance (AUROC) with different parameter settings}
      \label{fig:rf_tuning}

\begin{tikzpicture}

\begin{axis}[
title={MLlib Random Forest - tuning},
xlabel={number of trees},
ylabel={AUROC},
    y tick label style={
        /pgf/number format/.cd,
            fixed,
            fixed zerofill,
            precision=3,
        /tikz/.cd
    },
xmin=5, xmax=165,
ymin=0.634, ymax=0.66,
tick align=outside,
xmajorgrids,
x grid style={lightgray!92.026143790849673!black},
ymajorgrids,
y grid style={lightgray!92.026143790849673!black},
legend style={draw=white!80.0!black},
legend cell align={left},
legend entries={{depth = 4},{depth = 8},{depth = 16}}
]
\addplot [semithick, green!50.0!black, mark=triangle*, mark size=3, mark options={solid}]
table {%
10 0.635861897738276
20 0.636312176766189
30 0.636977880286977
40 0.638734446660559
50 0.637978516857033
60 0.638820623810479
70 0.639205800275521
80 0.63986689705606
90 0.641194945184474
100 0.643092808067015
110 0.643154311978396
120 0.64426679747348
130 0.642461721954148
140 0.642500538692224
150 0.643621276446228
160 0.644236428884419
};
\addplot [semithick, blue, mark=square, mark size=3, mark options={solid}]
table {%
10 0.645698414525277
20 0.649825406718019
30 0.65123824167222
40 0.652432994060624
50 0.651716948545089
};
\addplot [semithick, red, mark=pentagon*, mark size=3, mark options={solid}]
table {%
10 0.657010335445874
};
\end{axis}

\end{tikzpicture}
      \end{figure}
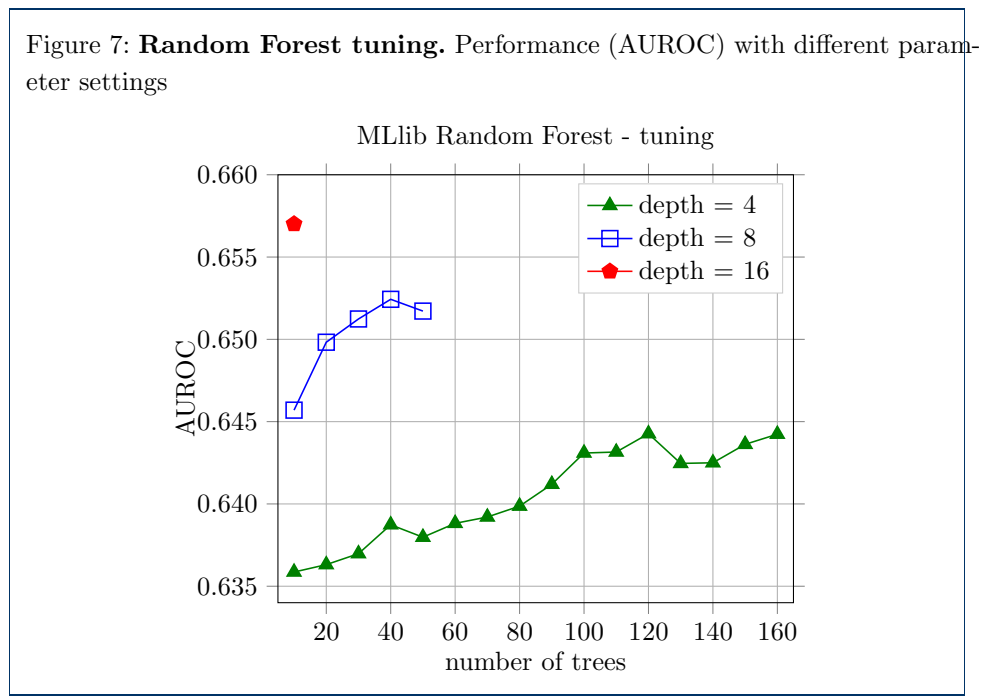
Figure \ref{fig:rf_tuning} shows the results.
With depth 4, the quality of the classification improves steadily until reaching a plateau after 100 trees.
The execution with 170 trees repeatedly failed, raising an \texttt{OutOfMemoryError} on our testbed.
With depth 8 and 10 trees, the quality improves of a not negligible point over the shallower version, and the gap still augments with more trees.
Unfortunately, the \texttt{OutOfMemoryError} appears even faster, with only 60 trees.
Finally, the only execution attempted with depth 16 scored 65.7\%. 
This result was obtained in an experiment lasting more than 17 hours, which would become, assuming linear scalability, more than 100 days for the tests with the complete dataset.
Similarly, building any forest with depth 8 has an unfeasible expected duration.
The solutions we tested on the whole dataset were thus focused on the forests with depth 4.

\subsection*{Experimental validation of a single-instance \CARgrowth}

In order to compare our work with previous works, we have evaluated a local, single-model version of  \BAC~on a number of medium-size datasets from the UCI repository, on which results for other associative classifiers were available.
The experiments showed that \BAC~performs similarly to CBA \cite{cba:ma1998integrating}, reaching higher accuracies as often as not.
Moreover, \BAC~reaches these results with a significantly lower number of rules, without any posterior pruning.
This sets the single-model \BAC~as a good choice for a base model in an ensemble, where usually shallow models are preferred as baseline models.

\section*{Related work}
\label{sec:rw}
Associative classifiers exist in a number of fashions, and a precise taxonomy has been already made in \cite{THABTAH2007AMining}.
Among all, we can distinguish classifiers exploiting CARs (Class Association Rules), as introduced in \cite{cba:ma1998integrating}, and others exploiting EPs (Emerging Patterns), like \cite{Dong1999:caep}. Our approach falls in the first category, together with works like \cite{Chen2006AMining,li2001cmar,yin2003cpar,Baralis2008AClassification,wang2005harmony,thabtah2005mcar,thabtah2004mmac,bechini2016mapreduce:mrac,zaiane2002classifying,Azevedo2001AnConviction,Venturini2016BAC:Frameworks}.
Since its introduction in \cite{cba:ma1998integrating}, the database coverage technique has been exploited with success by many classifiers, e.g. \cite{li2001cmar,Baralis2008AClassification,thabtah2005mcar,thabtah2004mmac}, and several others have also exploited similar techniques, e.g. \cite{xu2004caar,Dong1999:caep,li2001cmar}.
One of these is the redundant rule pruning, which scans again the set of rules found to delete the extensions of a rule that follow the rule itself, and that therefore are never applied \cite{li2001cmar}.
These techniques have proved to be very effective in the reduction of the model and the improvement of the quality of the classifier.
However, the amount of rules that are first extracted and then reordered is often enormous, demanding proportionate resources both in terms of memory and CPU.
We argue that, in order to scale to very large dimensions and effectively exploit the potentials of a MapReduce-like framework, an effective associative classifier should aim at reducing, if not eliminating, the contribution of these techniques to the reduction of the model size, and focus on the extraction of a small, good quality  subset of the rules.

Attempts to bring the training of an associative classifier onto a framework for parallel computing and scale to large datasets have been done in \cite{bechini2016mapreduce:mrac, Venturini2016BAC:Frameworks}.
The authors of \cite{bechini2016mapreduce:mrac} proposed a MapReduce solution based on a parallel implementation of FP-growth \cite{li2008pfp}, modified to extract CARs, followed by two pruning phases that are slight variants of the database coverage and the above-mentioned rule pruning. 
This solution was implemented and tested in the Hadoop framework.
The model proposed by \cite{Venturini2016BAC:Frameworks} is instead an ensemble of associative classifiers.
In this solution, many associative classifiers train their models in parallel on different samples of the original dataset, thus exploiting bagging.
Each one of the associative classifiers exploits FP-growth to generate CARs and the database coverage for pruning. 
The implementation runs on Apache Spark.
Both works follow the strategy of generating the complete set of CARs (for the portion of the dataset seen, in \cite{Venturini2016BAC:Frameworks}) and prune in a second phase.
While the general structure of our framework and the use of bagging is similar to \cite{Venturini2016BAC:Frameworks}, we addressed its main limitation, namely the large amount of memory used by the storage of the CARs extracted and not yet pruned.

We could not attempt a direct comparison with \cite{bechini2016mapreduce:mrac} as their code is not publicly available.
Furthermore, most of the datasets used in their experiments are characterized by continuous features.
Thus, the application domain is much different from the one of \BAC, that is designed to work on large-scale and large-domain categorical datasets.
The code from \cite{Venturini2016BAC:Frameworks} is instead open-source and publicly available\footnote{https://gitlab.com/ontic/bac}, and we attempted a direct comparison with \BAC~on the Criteo dataset.
Unfortunately, this algorithm cannot cope with such a very large domain, and runs out of memory in our testbed even with reduced samples of the dataset (a single day of logs).

Several works have already explored the possibility of combining more than one rule for prediction, thus defining weights akin to a score for each class.
The authors of \cite{li2001cmar} have proposed to use the top $K$ rules that match and weigh their vote with a weighted $\chi^2$-analysis.
In \cite{yin2003cpar}, the top rule for each class is first determined, then the prediction is made on the one that maximizes the Laplace accuracy.
\cite{Azevedo2001AnConviction} has proposed a weighted-voting based on some metrics, e.g. support, confidence and conviction of the rule.
\cite{wang2005harmony} sets as score the sum of the confidences for the matching rules.
All these techniques have been used selecting as label the class that maximizes the defined score.
The majority of associative classifiers, though, does not use a score and predicts the label with the first rule that matches the record \cite{Chen2006AMining,Baralis2008AClassification,thabtah2005mcar,thabtah2004mmac,bechini2016mapreduce:mrac}.

\section*{Conclusion}
\label{sec:conclusion}
In this work, we have proposed a technique to scale an associative classifier on very large datasets, namely a \BACDef~(\BAC).
We considered an in-memory cluster-computing architecture, Apache Spark.
In this architecture, the large availability of memory is heavily exploited to streamline the computation, avoiding disk access whenever possible, allowing an extremely faster sequential processing and caching of the intermediate results.
This scheme, and the available memory, have of course their limits.
In preliminary experiments, we identified the major issue of designing an associative classifier in this framework in the large number of extracted rules, which are only eventually pruned in the training phase.
Therefore, we have anticipated all the pruning into a novel extraction algorithm, \CARgrowth.
\BAC~trains an ensemble model by means of bagging, which eases the distribution of the computation. 
Each model is generated by an instance of \CARgrowth.
A final consolidation phase for the models of the ensemble and a new voting strategy help further reduce the size of the model and improve the quality of the predictions.

To validate our approach, we have performed experiments in a real large-scale scenario, a binary-labeled dataset with more than 4 billion records, 800 million distinct values in its categorical features and larger than 1.2 TB in storage.
The pruning done in \CARgrowth~has proved to be effective.
When executing database coverage pruning as a final step, a negligible fraction of rules are pruned by this technique, always below 5\%, without improvements in quality.
\BAC~demonstrated better performance than a state-of-the-art technique, a Random Forest, both in terms of quality of the prediction and execution time.
The best setting for \BAC~improves the AUROC upon the best for the Random Forest by 1\% with a total training time that is 25 times smaller.

The \BAC~classifier, differently from a Random Forest, generates a readable model.
The ``hashing trick'', which allows the Random Forest to deal with a large number of distinct values in the categoric fields, has the major drawback of making the model unintelligible by a human.
This hampers the usability of the model for decision-making and makes also extremely difficult its debugging.
\BAC~did not require hashing, though larger scales, i.e. billions or trillion of distinct values, might make it necessary.
In this scenario, the model produced by \BAC, without hashing, is made of rules containing the items exactly as they appear in the dataset, with all their semantics left intact.
We believe this feature to be highly valuable for a classification model.

Future works will experiment different model generation strategies. 
For example, we will introduce a projection by column in the ensemble like the one implemented in Random Forests.

\begin{backmatter}

\section*{Availability of data and materials}
The dataset supporting the conclusions of this article is available in the Criteo repository, http://labs.criteo.com/2013/12/download-terabyte-click-logs/

\section*{Competing interests}
  The authors declare that they have no competing interests.

\section*{Authors' contributions}
Luca Venturini conceived the algorithm, carried out the implementation and the experiments and drafted the manuscript with input from all authors. 
Elena Baralis and Paolo Garza provided reviews on the manuscript. 
All authors read and approved the final manuscript.

\section*{Acknowledgements}
The authors are thankful to Ernesto Valentino and Giulia Vasciaveo for having contributed to the code of \BAC.

The research leading to these results has received funding from the European Union’s Horizon 2020 research and innovation programme under grant agreement No 700256 (``I-REACT'' project). 

\bibliographystyle{vancouver} %
\bibliography{bmc_article.bib} 

\begin{thebibliography}{10}

\bibitem{el2013comparison}
El~Houby EM, Hassan MS.
\newblock Comparison between Associative Classification and Decision Tree for
  HCV Treatment Response Prediction.
\newblock World Academy of Science, Engineering and Technology, International
  Journal of Medical, Health, Biomedical, Bioengineering and Pharmaceutical
  Engineering. 2013;7(11):714--718.

\bibitem{pulvirentisurvey}
Apiletti D, Baralis E, Cerquitelli T, Garza P, Pulvirenti F, Venturini L.
\newblock Frequent Itemsets Mining for Big Data: A Comparative Analysis.
\newblock Big Data Research. 2017;.

\bibitem{THABTAH2007AMining}
Thabtah F.
\newblock A review of associative classification mining.
\newblock The Knowledge Engineering Review. 2007;22(01):37--65.

\bibitem{bechini2016mapreduce:mrac}
Bechini A, Marcelloni F, Segatori A.
\newblock A MapReduce solution for associative classification of big data.
\newblock Information Sciences. 2016;332:33--55.

\bibitem{Venturini2016BAC:Frameworks}
Venturini L, Garza P, Apiletti D.
\newblock BAC: A bagged associative classifier for big data frameworks.
\newblock In: East European Conference on Advances in Databases and Information
  Systems. Springer; 2016. p. 137--146.

\bibitem{cba:ma1998integrating}
Liu B, Hsu W, Ma Y.
\newblock Integrating classification and association rule mining.
\newblock In: Proceedings of the Fourth International Conference on Knowledge
  Discovery and Data Mining. AAAI Press; 1998. p. 80--86.

\bibitem{breiman1996some}
Breiman L.
\newblock Some properties of splitting criteria.
\newblock Machine Learning. 1996;24(1):41--47.

\bibitem{li2001cmar}
Li W, Han J, Pei J.
\newblock CMAR: Accurate and efficient classification based on multiple
  class-association rules.
\newblock In: Data Mining, 2001. ICDM 2001, Proceedings IEEE International
  Conference on. IEEE; 2001. p. 369--376.

\bibitem{Baralis2008AClassification}
Baralis E, Chiusano S, Garza P.
\newblock A lazy approach to associative classification.
\newblock IEEE Transactions on Knowledge and Data Engineering.
  2008;20(2):156--171.

\bibitem{mllib2016meng}
Meng X, Bradley J, Yavuz B, Sparks E, Venkataraman S, Liu D, et~al.
\newblock MLlib: Machine Learning in Apache Spark.
\newblock J Mach Learn Res. 2016 Jan;17(1):1235--1241.

\bibitem{landset2015survey}
Landset S, Khoshgoftaar TM, Richter AN, Hasanin T.
\newblock A survey of open source tools for machine learning with big data in
  the Hadoop ecosystem.
\newblock Journal of Big Data. 2015;2(1):24.

\bibitem{singh2015survey}
Singh D, Reddy CK.
\newblock A survey on platforms for big data analytics.
\newblock Journal of Big Data. 2015;2(1):8.

\bibitem{Chapelle2014SimpleAdvertising}
Chapelle O, Manavoglu E, Rosales R.
\newblock Simple and scalable response prediction for display advertising.
\newblock ACM Transactions on Intelligent Systems and Technology (TIST).
  2015;5(4):61.

\bibitem{chen2016xgboost}
Chen T, Guestrin C.
\newblock Xgboost: A scalable tree boosting system.
\newblock In: Proceedings of the 22Nd ACM SIGKDD International Conference on
  Knowledge Discovery and Data Mining. ACM; 2016. p. 785--794.

\bibitem{WeinbergerFeatureLearning}
Weinberger K, Dasgupta A, Langford J, Smola A, Attenberg J.
\newblock Feature hashing for large scale multitask learning.
\newblock In: Proceedings of the 26th Annual International Conference on
  Machine Learning. ACM; 2009. p. 1113--1120.

\bibitem{scikit-learn}
Pedregosa F, Varoquaux G, Gramfort A, Michel V, Thirion B, Grisel O, et~al.
\newblock Scikit-learn: Machine Learning in {P}ython.
\newblock Journal of Machine Learning Research. 2011;12:2825--2830.

\bibitem{r-randomforest}
Liaw A, Wiener M.
\newblock Classification and Regression by randomForest.
\newblock R News. 2002;2(3):18--22.

\bibitem{witten2016data}
Witten IH, Frank E, Hall MA, Pal CJ.
\newblock Data Mining: Practical machine learning tools and techniques.
\newblock Morgan Kaufmann; 2016.

\bibitem{bradley1997auroc}
Bradley AP.
\newblock The use of the area under the ROC curve in the evaluation of machine
  learning algorithms.
\newblock Pattern recognition. 1997;30(7):1145--1159.

\bibitem{Sebastiani:2002:MLA:505282.505283}
Sebastiani F.
\newblock Machine Learning in Automated Text Categorization.
\newblock ACM Comput Surv. 2002 Mar;34(1):1--47.

\bibitem{Dong1999:caep}
Dong G, Zhang X, Wong L, Li J.
\newblock In: Arikawa S, Furukawa K, editors. CAEP: Classification by
  Aggregating Emerging Patterns. Berlin, Heidelberg: Springer Berlin
  Heidelberg; 1999. p. 30--42.

\bibitem{Chen2006AMining}
Chen G, Liu H, Yu L, Wei Q, Zhang X.
\newblock A new approach to classification based on association rule mining.
\newblock Decision Support Systems. 2006;42(2):674--689.

\bibitem{yin2003cpar}
Yin X, Han J.
\newblock CPAR: Classification based on predictive association rules.
\newblock In: Proceedings of the 2003 SIAM International Conference on Data
  Mining. SIAM; 2003. p. 331--335.

\bibitem{wang2005harmony}
Wang J, Karypis G.
\newblock HARMONY: Efficiently mining the best rules for classification.
\newblock In: Proceedings of the 2005 SIAM International Conference on Data
  Mining. SIAM; 2005. p. 205--216.

\bibitem{thabtah2005mcar}
Thabtah F, Cowling P, Peng Y.
\newblock MCAR: multi-class classification based on association rule.
\newblock In: Computer Systems and Applications, 2005. The 3rd ACS/IEEE
  International Conference on. IEEE; 2005. p.~33.

\bibitem{thabtah2004mmac}
Thabtah FA, Cowling P, Peng Y.
\newblock MMAC: A new multi-class, multi-label associative classification
  approach.
\newblock In: Data Mining, 2004. ICDM'04. Fourth IEEE International Conference
  on. IEEE; 2004. p. 217--224.

\bibitem{zaiane2002classifying}
Za{\"\i}ane OR, Antonie ML.
\newblock Classifying text documents by associating terms with text categories.
\newblock In: Australian computer Science communications. vol.~24. Australian
  Computer Society, Inc.; 2002. p. 215--222.

\bibitem{Azevedo2001AnConviction}
Jorge AM, Azevedo PJ.
\newblock An experiment with association rules and classification: Post-bagging
  and conviction.
\newblock In: International Conference on Discovery Science. Springer; 2005. p.
  137--149.

\bibitem{xu2004caar}
Xu X, Han G, Min H.
\newblock A novel algorithm for associative classification of image blocks.
\newblock In: Computer and Information Technology, 2004. CIT'04. The Fourth
  International Conference on. IEEE; 2004. p. 46--51.

\bibitem{li2008pfp}
Li H, Wang Y, Zhang D, Zhang M, Chang EY.
\newblock Pfp: parallel fp-growth for query recommendation.
\newblock In: Proceedings of the 2008 ACM conference on Recommender systems.
  ACM; 2008. p. 107--114.

\end{thebibliography}

\end{backmatter}
\end{document}